\documentclass[journal]{IEEEtran}
\usepackage{graphicx}
\usepackage{multicol}
\usepackage{multirow}
\usepackage{color}
\usepackage{subcaption}
\usepackage{pifont}
\usepackage{amssymb}
\usepackage{amsmath}

\ifCLASSINFOpdf
\else
\fi
\begin{document}
%
\title{Degradation-Aware Self-Attention Based Transformer for Blind Image Super-Resolution}
%
%
%
        \author{Qingguo~Liu,~\IEEEmembership{}
                Pan Gao,~\IEEEmembership{}
                Kang Han,~\IEEEmembership{}
                Ningzhong Liu,~\IEEEmembership{}
                and~Wei Xiang~\IEEEmembership{}
        \thanks{Qingguo~Liu, Pan Gao, and Ningzhong Liu are with the College of Computer Science and Technology, Nanjing University of Aeronautics and Astronautics, Nanjing 211106, China, E-mail: {liuqingguo, Pan.Gao}@nuaa.edu.cn, liunz@163.com.}
        \thanks{Kang Han and Wei Xiang are with the School of Computing, Engineering and Mathematical Sciences at La Trobe University \{e-mail:K.Han, w.xiang@latrobe.edu.au\}.}}
\markboth{Journal of \LaTeX\ Class Files,~Vol.~14, No.~8, August~2015}%
{Shell \MakeLowercase{\textit{et al.}}: Bare Demo of IEEEtran.cls for IEEE Journals}
%



\maketitle

\begin{abstract}
    Compared to CNN-based methods, Transformer-based methods achieve impressive image restoration outcomes due to their abilities to model remote dependencies. However, how to apply Transformer-based methods to the field of blind super-resolution (SR) and further make an SR network adaptive to degradation information is still an open problem. In this paper, we propose a new degradation-aware self-attention-based Transformer model, where we incorporate contrastive learning into the Transformer network for learning the degradation representations of input images with unknown noise. In particular, we integrate both CNN and Transformer components into the SR network, where we first use the CNN modulated by the degradation information to extract local features, and then employ the degradation-aware Transformer to extract global semantic features. We apply our proposed model to several popular large-scale benchmark datasets for testing, and achieve the state-of-the-art performance compared to existing methods. In particular, our method yields a PSNR of 32.43 dB on the Urban100 dataset at $\times$2 scale,  0.94 dB higher than  DASR, and 26.62 dB on the Urban100 dataset at $\times$4 scale,  0.26 dB improvement over KDSR, setting a new benchmark in this area. Source code is available at: https://github.com/I2-Multimedia-Lab/DSAT/tree/main.
\end{abstract}


%
\IEEEpeerreviewmaketitle

\section{Introduction}
%
%
%
%

\IEEEPARstart{I}{mage} super-resolution (SR) aims to reconstruct high-resolution (HR) images from low-resolution (LR) images. Convolutional neural networks (CNNs) have been the mainstay in the field of image SR reconstruction in recent years. This paper~\cite{8723565} reviews numerous CNN-based SR methods. The majority of these methods are non-blind and require designing complex and very deep network structures \cite{li2019feedback,fritsche2019frequency,zhang2021plug,zhang2021designing,kim2016accurate}. However, in these customized CNN structures, the convolution kernels are content independent of each other, which fails to capture long-range pixel dependencies and thus usually does not bring about the best results. In addition, most of these methods are based on the assumption that the degradation of the original image is known and fixed. However, the degradation in real-world applications is often complex and unknown. Therefore, these methods suffer from severe performance degradation when the actual degradation differs from their assumptions. In order to reconstruct images with unknown degradation in real-world applications, a degradation estimate is needed to provide degradation information for non-blind SR networks. However, unexpectedly, most non-blind SR methods are sensitive to degradation estimates. As a result, SR networks can further amplify errors in the degradation estimates and thus produce unsatisfactory results. 

To address the above challenge, a plethora of blind super-resolution reconstruction methods have been developed. At present, blind SR image reconstruction methods include mainly explicit kernel estimation \cite{gu2019blind,luo2022deep}, and implicit degradation representation \cite{wang2021unsupervised,xia2022knowledge}.  Gu \emph{et al.} proposed an iterative kernel correction (IKC) method~\cite{gu2019blind} through iteratively correcting the estimated degradation by taking into account the previous SR results. In this way, satisfactory results can be produced gradually. However, the IKC method requires several iterations during testing, which is time-consuming.

\begin{figure}
	\centering
	\includegraphics[width=1.0\linewidth]{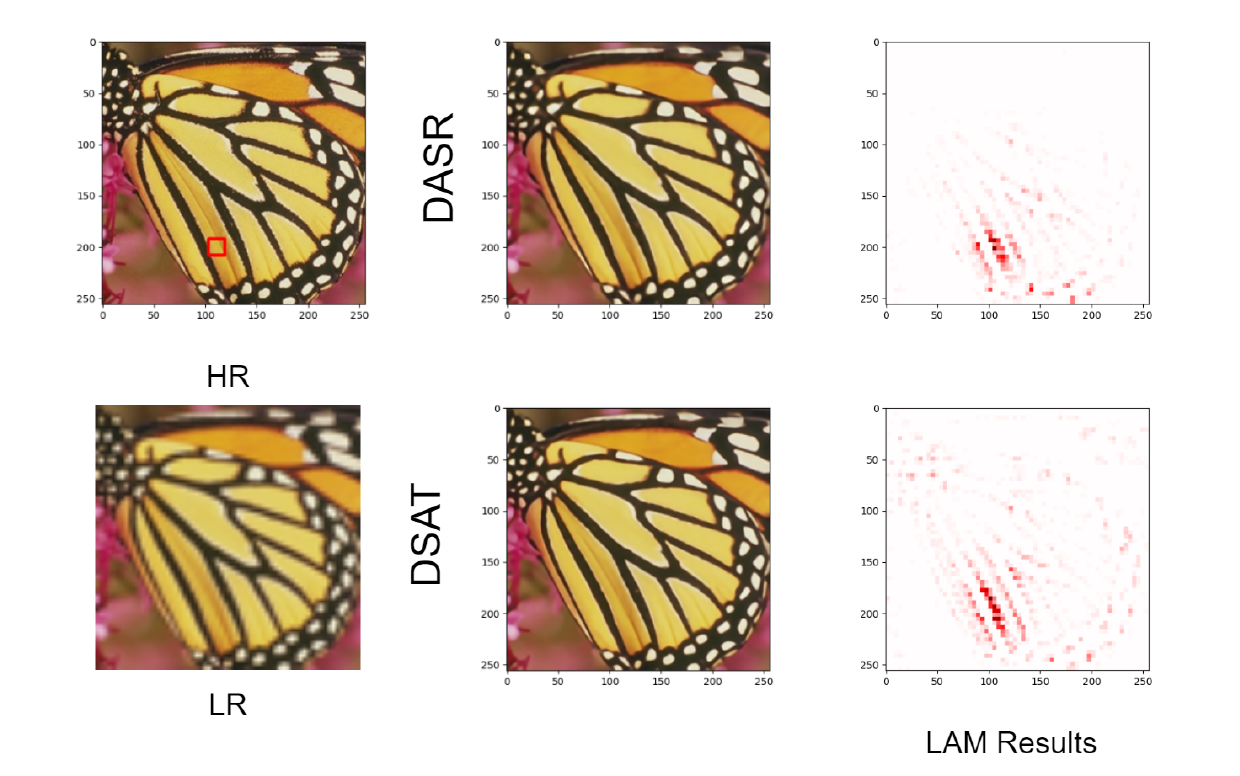}
	
	\caption{Qualitative comparison of the LAM results between the DASR and proposed DSAT.}
	\label{fig:lam}
\end{figure}

The state-of-the-art kernel estimation method, i.e., DCLS \cite{luo2022deep}, also relies very much on the estimation of fuzzy kernels. If the estimated fuzzy kernel is incorrect, it will lead to a fatal error in the final result. Furthermore, this explicit kernel estimation is designed mainly for fuzzy kernel estimation, but the real degradation also includes other degradation factors such as noise. On the other hand, the degradation-based learning method KDSR \cite{xia2022knowledge} uses knowledge distillation to estimate degradation, which may not be stable because it is difficult to guarantee that the student network learns the same degradation extraction ability as the teacher network in practice. As for DASR \cite{wang2021unsupervised}, which is another pure CNN blind SR method for degradation-based learning, due to the limitations of CNNs, DASR does not make full use of input pixels. Fig.~\ref{fig:lam} shows the attribution analysis method LAM \cite{gu2021interpreting} result and demonstrates that our model DSAT can make better use of more pixels.

Meanwhile, we observe that the Swin Transformer \cite{liu2021swin} can use the self-attention mechanism to capture global interactions between context. On the one hand, it enjoys the same advantage of CNNs in processing large-sized images due to the local attention mechanism. On the other hand, it has the advantage of the Transformer networks to model remote dependencies using a shifted window scheme. Inspired by this, Liang \emph{et al.} \cite{liang2021swinir} proposed a Swin Transformer-based image recovery model, but it is generally a non-blind image SR method, which cannot cope with complex degradation in practical applications.
\begin{figure*}[t]
	\vspace{-0.5em}
	\centering
	\includegraphics[width=1.0\linewidth]{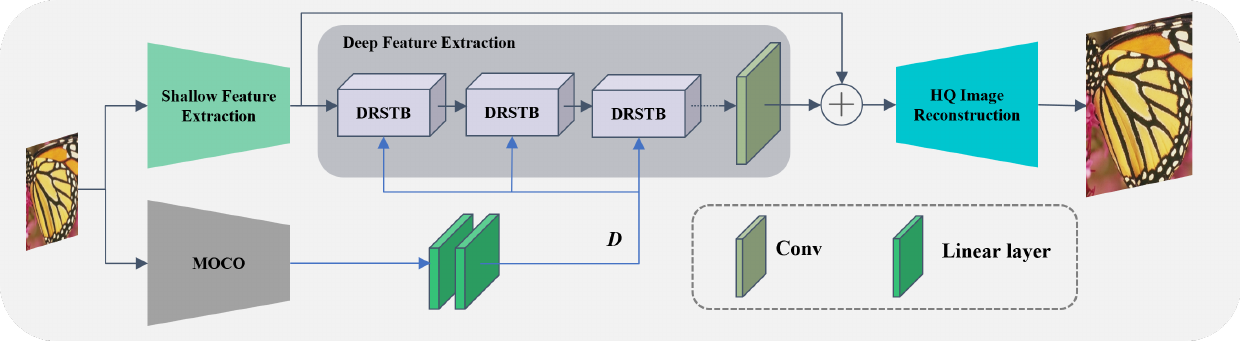}
	
	\caption{End-to-end architecture of the proposed DSAT for image super-resolution.}
	\label{fig:onecol}
	\vspace{-2mm}
\end{figure*}

\begin{figure*}
	\flushleft
	\begin{subfigure}{0.28\linewidth}
		\centering
		\includegraphics[height=200pt]{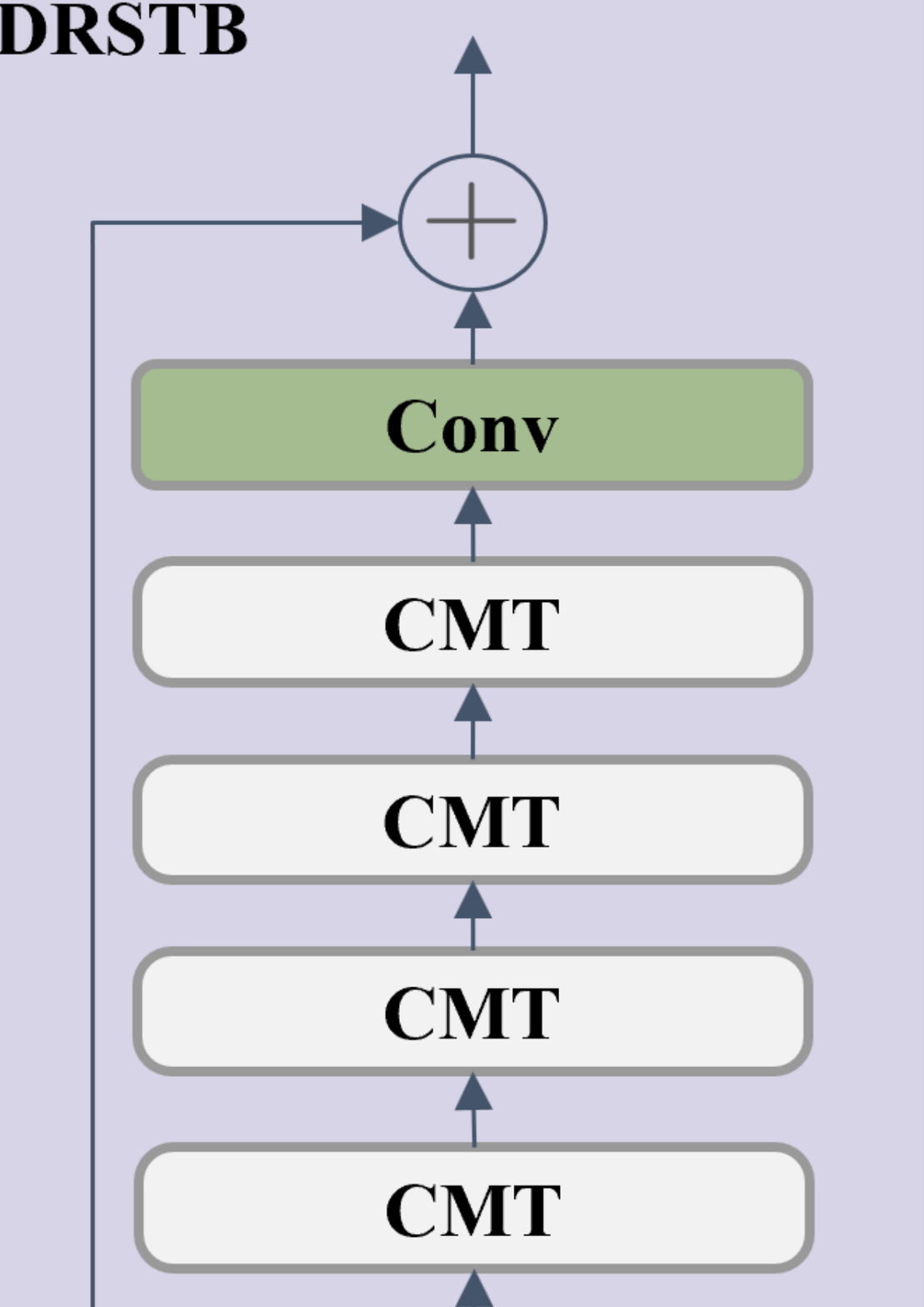}
		\caption{DRSTB.}
		\label{fig:short2-a}
	\end{subfigure}
	\begin{subfigure}{0.25\linewidth}
		\includegraphics[height=200pt]{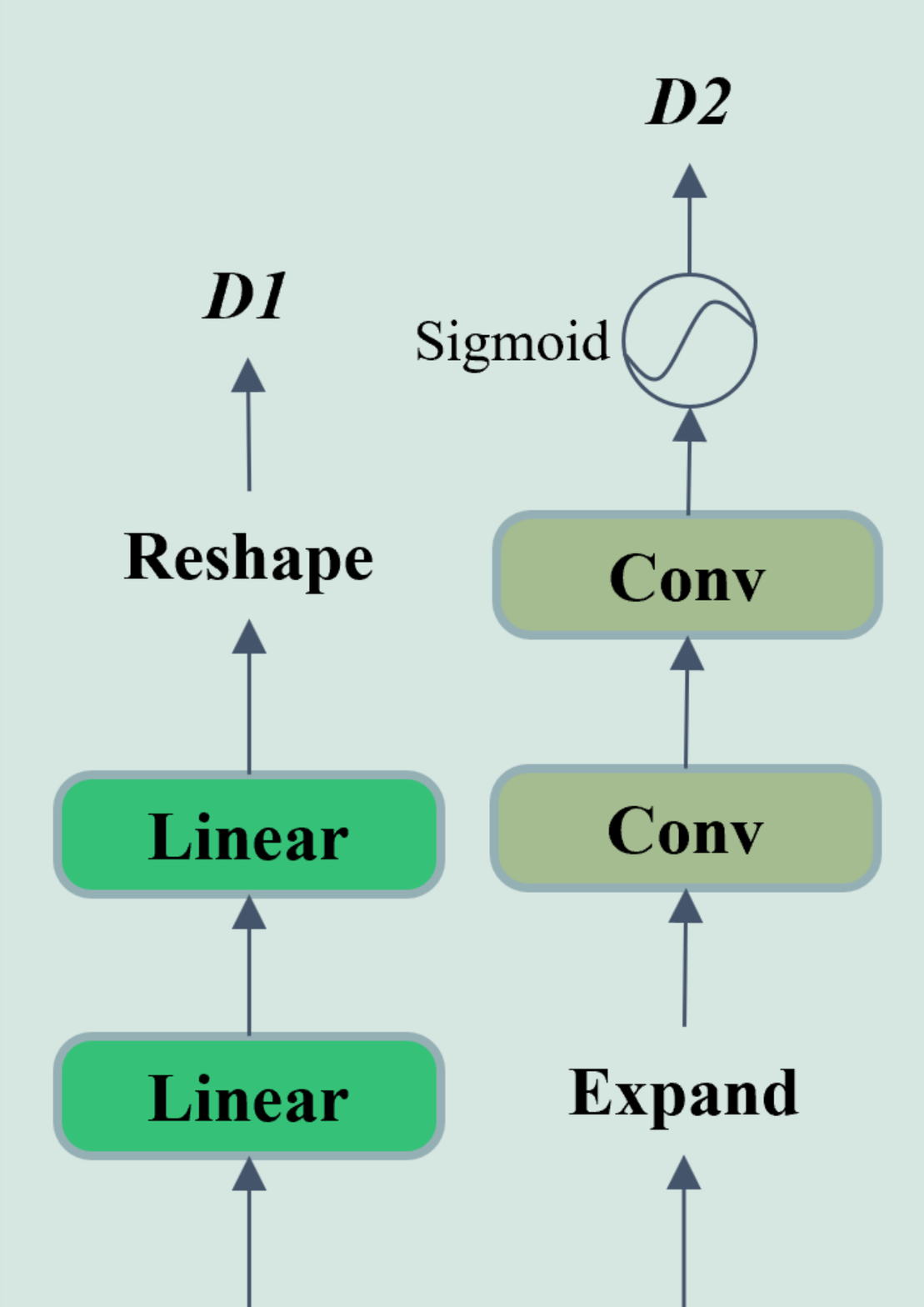}
		\caption{Generate D1 and D2.}
		\label{fig:short2-b}
	\end{subfigure}
	\begin{subfigure}{0.4\linewidth}
		\includegraphics[height=200pt]{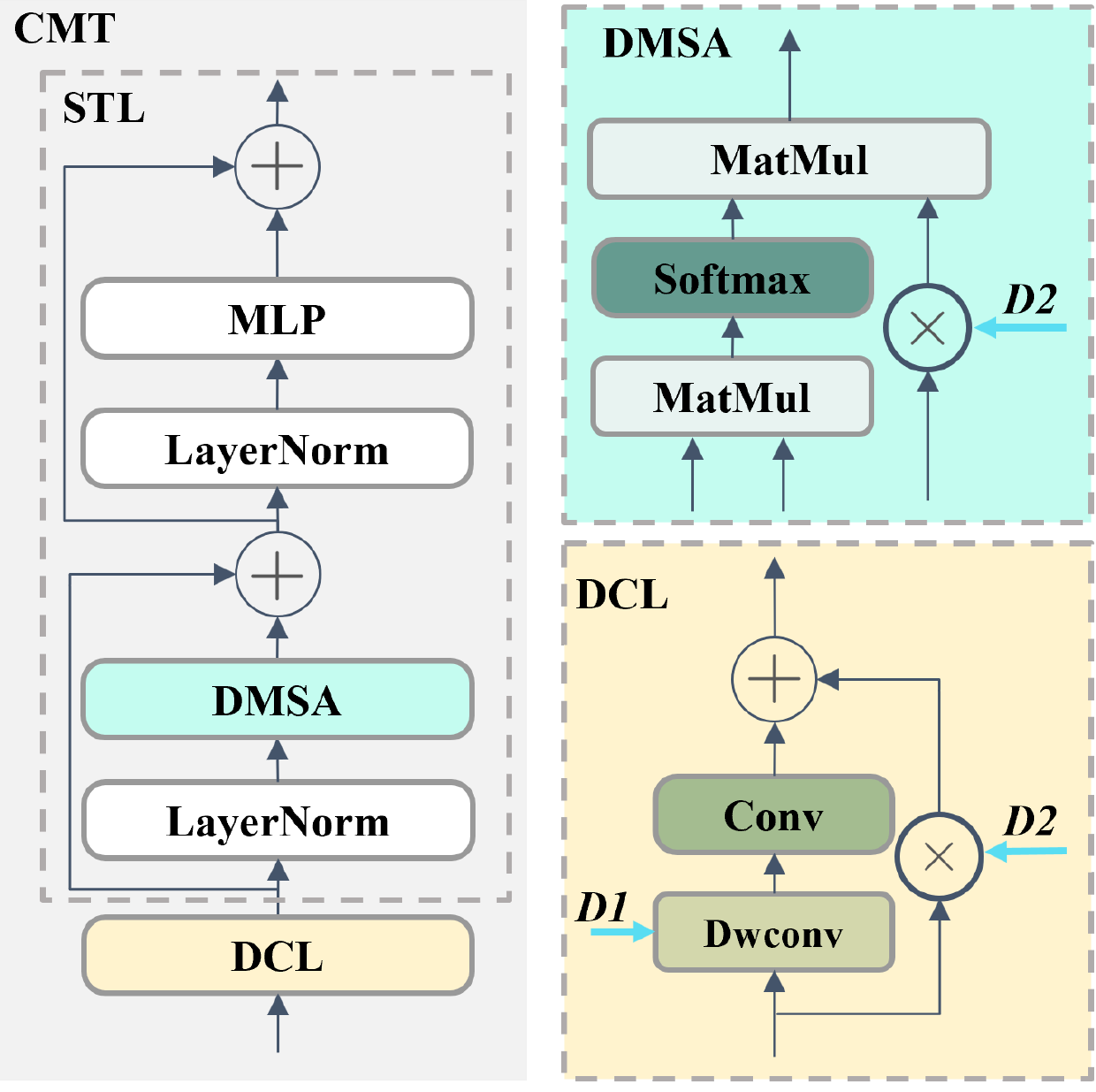}
		\caption{CMT.}
		\label{fig:short2-c}
	\end{subfigure}
	\caption{Main components of the proposed network. $\rm{D}$ represents the learned degradation representation.}
	\label{fig:short2}
\end{figure*}

Different from the above methods, in this paper, we propose a  Degradation-aware Self-Attention Transformer for blind image SR, dubbed the DSAT, which includes a degradation learning module, a shallow feature extraction module, a deep feature extraction module, and a high-quality reconstruction module. We introduce degradation representation information to the deep feature extraction module to distinguish potential degradation from other degradation by use of contrastive learning to complement the SR network for a more accurate reconstruction. Specifically, the degradation module uses the MoCo \cite{he2020momentum} method to learn potential representations, where image patches from the same image will obtain the same degradation representation. The deep feature extraction module is mainly composed of the Degradation-aware Residual Swin Transformer blocks (DRSTB), each of which consists of several CNN Mixed Transfomers (CMTs) and a convolutional layer. For each CMT, it consists of a Degradation-aware CNN layer (DCL) and a Swin Transformer Layer (STL). The DCL employs depth-wise convolution and channel attention fusion blocks to integrate degradation perceptual information to extract local features. Meanwhile, to enable the Transformer to better extract global features from the degraded input image, we design a new attention mechanism in explicit consideration of degradation perceptual information. Finally, shallow and deep features are fused into the reconstruction module to achieve high-quality image SR reconstruction.

Compared with the above popular CNN- and Swin Transformer-based image SR models, our Transformer-based blind image SR method has two advantages. Firstly, we design a CNN Mixed Transformer module incorporating degradation information for feature extraction, in an effort to make full use of the fact that the CNN can exploit the inductive bias to compensate for the shortcoming that Transformers struggle to acquire local finer-grained details. Secondly, we design an attention mechanism to fuse potential degradation representation information, which can make the Transformer adaptive to unknown degradation during learning. Experimental results show that our network can handle various types of degradation and produce promising results for both synthetic and real images under blind settings. To the best of the authors' knowledge, our work is the first to apply the Transformer to the field of blind image SR reconstruction and achieve the SOTA performance.

\section{Related Work}
\label{sec:intro} 
\subsection{Single Image Super-Resolution}
Single-image SR is a classical problem in computer vision, which aims to reconstruct an HR image from a single LR image. The main challenge lies in the fact that finding an inherently unique solution is elusive due to multiple potential solutions for a given low-resolution pixel. In fact, the information loss from HR to LR can be partially recovered by leveraging the prior information in the image. It is usually observed that SR methods contain three main modules, i.e., shallow feature extraction, deep feature extraction or mapping based on the extracted shallow features, and SR output reconstruction. In recent years, many improvements have been made to the deep feature extraction and SR reconstruction modules, such as introducing residual blocks \cite{wang2018esrgan,lim2017enhanced,tong2017image}, recursive or recurrent structures \cite{kim2016deeply,tai2017image}, attention mechanisms \cite{zhang2018image,dai2019second}, sub-pixel convolution \cite{shi2016real}, reference-based \cite{9868165}, latent style guided \cite{10145603}, etc. These techniques have made significant progress in reconstruction accuracy and efficiency, so that non-blind SISR under bicubic downsampling has almost reached maturity. However, these non-blind models often struggle to effectively generalize to input images that exhibit complexities beyond the specific degradation patterns they are designed to handle.


\subsection{Vision Transformer}

Recently, Transformer models have been widely used in a variety of computer vision applications, such as object detection \cite{carion2020end,liu2020deep,liu2021swin}, segmentation \cite{liu2021swin,wu2020visual,zheng2021rethinking,cao2021swin}, and crowd counting \cite{liang2022transcrowd,sun2021boosting}. Due to their excellent performance, Transformers have also been introduced into the field of single-image SR \cite{chen2021pre,liang2021swinir,wang2022general,cao2021video,10207832}. The two representative methods are the SwinIR \cite{liang2021swinir} and IPT \cite{chen2021pre}. The SwinIR introduces the Swin Transformer to non-blind SR for the first time and achieves the best results in classical image SR reconstruction, image denoising, and compression artifact removal. The IPT, on the other hand, is pre-trained on the ImageNet dataset to further exploit the potential of Transformers in SR. Although the SwinIR and IPT have shown superior performances on images with known degradation, they still struggle to deal with complex and unknown degradation. On the contrary, our method with degradation learning is adaptive to complex and unknown degradation.

\subsection{Contrastive Learning}

Contrastive learning \cite{tian2020contrastive,he2020momentum,chen2020simple} has been widely used in unsupervised representation learning. Previous methods typically perform representation learning by minimizing the difference between the output and the target. Rather than using predefined targets, contrastive learning maximizes mutual information in the representation space. Specifically, the representation of a query sample should attract those that are similar to itself and reject those that are otherwise. 
MoCo is a classical algorithm in contrast learning. MoCo can learn the similarity scores between distinct features in an unsupervised way, and this idea is well suited to meet the needs of blind image SR reconstruction that cannot directly obtain information about image degradation. Based on the idea of MoCo and contrastive learning, we can make similar degraded features as clustered as possible so that we can find the representations of various types of degradation. Similarly, we can use this representation to empower the network to tackle more complex degradation.
In this paper, the image patches generated with the same degradation are considered as positive correlation \cite{bell2019blind,zhang2020deep}, and contrastive learning is performed to obtain content-invariant degradation representations as a \textit{priori} information.



\section{Method}

Our approach stems from our finding that pure CNNs produce unclear or even incorrect texture details when processing images with complex degradation, which is clearly not the result we want. Therefore, we want to implicitly generate a degradation representation that can distinguish different types of degradation by contrastive learning, and then use this degradation representation to guide the Transformer network to process complex unknown types of degradation, which enhances the representation learning ability of the network.
In this paper, we propose a blind image SR with unsupervised degradation representation learning based on the Swin Transformer. 

	


	\subsection{Network Architecture}
	As shown in Fig. \ref{fig:onecol}, our method consists of four modules, namely the degradation learning module, shallow feature extraction module, deep feature extraction module, and high-quality image reconstruction module.
	
	\textbf{Degradation Representation Learning.}
	Our representation learning aims to achieve implicit extraction of degradation information from $LR$ images in an unsupervised manner. More specifically, one image patch is randomly selected from an $LR$ image as a query patch, other patches in the same $LR$ image as positive samples, and patches from other $LR$ images as negative samples. Then, we feed the obtained encoding queue into MoCo to implement degradation encoding learning to obtain $x$, $x^{+}$ and $x^{-}$, such that $x$ is close to $x^{+}$ and far from $x^{-}$.
	
	\textbf{Shallow Feature Extraction.}
	Given a low-quality (LQ) input $I_{LQ} \in \mathbb{R}^{H \times W \times C_{in}}$($H$, $W$ and $C_{in}$ are the image height, width and number of input channels, respectively), we use a $ 3 \times 3 $ convolutional layer $H_{SF}(\cdot)$ to extract shallow feature $F_{0} \in {R}^{H \times W \times C}$ as follows
	\begin{equation} 
		F_{0} = H_{SF}(I_{LQ})
	\end{equation}
	where $C$ is the number of feature channels, $I_{LQ}$ is the LR image, $H_{SF}(\cdot)$ is the shallow feature extraction module. 
	
	\textbf{Deep Feature Extraction.}
	We use $F_{0}$ after shallow feature extraction as the input to the deep feature extraction module as follows
	\begin{equation} 
		F_{DF}=H_{DF}(F_{0})
	\end{equation}
	where $H_{DF}(\cdot)$ is the deep feature extraction module and contains $K$ Degradation-aware Residual Swin Transformer Blocks (DRSTB) and a $ 3 \times 3 $ convolutional layer. More specifically, intermediate features $F_{1}, F_{2}, . . . , F_{K}$ and the output deep feature $F_{DF}$ are extracted block by block as follows
	\begin{equation} 
		F_{i} = H_{DRSTB_{i}(F_{i-1})}, i = 1, 2, . . . , K
	\end{equation}
	\begin{equation} 
		F_{DF} = H_{CONV}(F_{K})
	\end{equation}
	where $H_{DRSTB_{i}(\cdot)} $ denotes the $i\rm{th}$ DRSTB and $H_{CONV}$ is the last convolutional layer.
	
	\textbf{Image Reconstruction.}
	We reconstruct the high-quality image $I_{SR}$ by aggregating shallow and deep features as follows
	\begin{equation} 
		I_{SR} = H_{REC}(F_{0} + F_{DF})
	\end{equation}
	where $H_{REC}(\cdot)$ is the high-quality image reconstruction module.
	
	\textbf{Loss Function.}
	We design two loss functions, i.e., the $L_{1}$ pixel loss and the degradation perceptual loss ${L_{degrad}}$. Here contrastive learning is implemented using the InfoNCE \cite{dyer2014notes} loss. To obtain better degradation-aware information, we need one queue that contains samples with various contents and types of degradation. Therefore, we first randomly select $B$ LR images (representing $B$ different types of degradation) and then randomly crop two patches from each image. These patches are then encoded into ${p_{i}^{1} ,p_{i}^{2} \in \mathbb{R}^{256}},$ where $p_{i}^{1}$ is the embedding of the first patch of the $i\rm{th}$ image and $p_{i}^{2}$ is the embedding of the second patch of the $i\rm{th}$ image. For the $i\rm{th}$ image, $p_{i}^{1}$ and $p_{i}^{2}$ are referred to as the query and the positive sample, respectively. The degradation loss is defined as
	
	\begin{equation} 
		L_{\rm{degrad}}=\sum_{i=1}^{B}-\log \frac{\exp \left(p_{i}^{1} p_{i}^{2} / \tau\right)}{\sum_{j=1}^{N_{\rm{queue}}} \exp \left(p_{i}^{1} p_{\rm{queue}}^{j} / \tau\right)}
	\end{equation}
	where $N_{\rm{queue}}$ is the number of samples in the queue and $P^{j}_{\rm{queue}}$ represents the $j\rm{th}$ negative sample.
	
	For the SR network, we use the $ L_{1}$ pixel loss as follows
	\begin{equation} 
		\mathcal{L_{SR}}=\left\|I_{S R}-I_{H Q}\right\|_{1}
	\end{equation}
	where $I_{S R}$ is obtained by taking $I_{LQ}$ as the input to our SR network, and $I_{HQ}$ is the corresponding ground-truth HQ image.
	
	The end-to-end loss of the whole network is defined as
	\begin{equation} 
		L_{loss} = L_{degrad } + \mathcal{L_{SR}}.
	\end{equation}
	
%

	\subsection{Degradation-aware Residual Swin Transformer Block}
	As shown in Fig. \ref{fig:short2-a}, the DRSTB consists of several CNN Mixed Transformer blocks (CMT) and a convolutional layer. Given the input feature $F_{i,0}$ of the $i\rm{th}$ DRSTB, we first obtain the degradation representation $F_{DA}$ by degradation learning block, and then extract features $F_{i,1}, F_{i,2}, . . . ,F_{i,L},$ by using $L$ CNN Mixed Transformer Blocks as follows
	\begin{equation} 
		F_{i, j}=H_{CMT_{i, j}}\left(F_{i, j-1}, F_{DA}\right), \quad j=1,2, \ldots, L
	\end{equation}
	where $H_{CMT_{i, j}}(\cdot)$ is the $j\rm{th}$ CMT block in the $i\rm{th}$ DRSTB.

	\textbf{CNN Mixed Transformer Block.}
	Each CMT consists of a Degradation-aware CNN Layer (DCL) and a Swin Transformer Layer (STL). As shown in Fig. \ref{fig:short2-c}, a DCL includes depth-wise convolution and channel weight adjustment, and the depth-wise convolution kernel and channel coefficients are derived using the degradation representation obtained earlier. Specifically, as shown in Fig. \ref{fig:short2-b}, the depth-wise convolution kernel parameters $D1 \in \mathbb{R}^{C \times 1 \times 3 \times 3}$ are obtained from the degradation representation $D$ by passing through two fully-connected layers and a reshape layer. Channel coefficients $D2 \in \mathbb{R}^{C \times H \times W}$ are obtained from the degradation representation $D$ by first dimensional augmentation through expansion and then passing through two $1 \times 1$ convolution layers and a sigmoid activation layer. In the DCL, we use $D1$ to control the kernel parameters of the depthwise convolution and $D2$ to generate modulation coefficients to rescale the components of the different channels in $F_{i, j-1}$. The STL is based on our modified multi-head self-attention of the original Transformer layer. The main improvement lies in that we use the learned degradation to modulate the channel attention in the self-attention calculation.
	As shown in Fig. \ref{fig:short2-c}, after the DCL, given an input of size $H \times W \times C$, the degradation-aware Swin Transformer layer first reshapes the input to a $\dfrac{HW}{M^{2}} \times M^{2} \times C$ feature by partitioning the input into non-overlapping $M \times M$ local windows, where $\dfrac{HW}{M^{2}}$ is the total number of windows. Then, it computes the local self-attention separately for each window. After the linear layer of $X \in \mathbb{R}^{M^{2} \times C}$, we also add weights based upon degradation representations to the value component to adjust the channel attention. To accommodate the attention weights, we change the dimension of $D2$ to $D2 \in \mathbb{R}^{1 \times d}$. 

\begin{table*}[t]
		\caption{Ablation study, where PSNR and SSIM on Set14 for different kernel widths and noises are shown.}
		\vspace{-3mm}
		\centering
		\label{table:1}
		\resizebox{\textwidth}{!}{
			\begin{tabular}{cccccccccccc}
				\hline
				\multirow{3}{*}{Method} & \multirow{3}{*}{Degradation Learning} & \multirow{3}{*}{DCL}  & \multirow{3}{*}{Attention Weights} & \multicolumn{8}{c}{$\sigma$/noise}  \\
				& & & & \multicolumn{2}{c}{2.4/0} &\multicolumn{2}{c}{2.4/10} & \multicolumn{2}{c}{3.6/5} & \multicolumn{2}{c}{3.6/15} \\
				& & & & PSNR $\uparrow$ & SSIM $\uparrow$ & PSNR $\uparrow$ & SSIM $\uparrow$ & PSNR $\uparrow$ & SSIM $\uparrow$ & PSNR $\uparrow$ & SSIM $\uparrow$ \\
				\hline
				model1 & \ding{56} & \ding{52} & \ding{52} & 27.822 & 0.7526 & 25.908 & 0.6699 & 25.607 & 0.6561 & 24.466 & 0.6147  \\
				model2 & \ding{52} & \ding{52} & \ding{56} & 27.925 & 0.7552 & 25.954 & 0.6707 & 25.677 & 0.6584 & 24.522 & 0.6168  \\
				model3 & \ding{52} & \ding{56} & \ding{52} & 28.063 & 0.7593 & 25.974 & 0.6722 & 25.673 & 0.6595 & 24.513 & 0.6178   \\
				model4 & \ding{52} & \ding{56} & \ding{56} & 28.058 & 0.7584 & 25.974 & 0.6719 & 25.654 & 0.6586 & 24.494  & 0.6169  \\
				model5 & \ding{52} & \ding{52} & \ding{52} & \textbf{28.085} & \textbf{0.7621} & \textbf{26.008} & \textbf{0.6738} & \textbf{25.715} & \textbf{0.6597} & \textbf{24.546} & \textbf{0.6188} \\
				\hline
				
		\end{tabular}}
	\end{table*}

\begin{figure*}[t]
    \centering
    \includegraphics[width=1.0\linewidth]{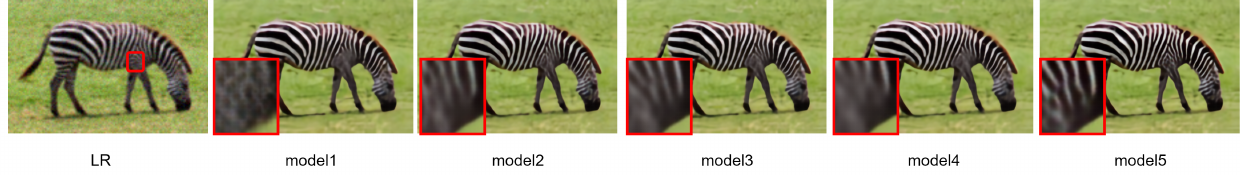}
    
    \caption{Visualization of the ablation experiments.}
    \label{fig:alb}
    \vspace{-2mm}
\end{figure*}

 \begin{table*}
	\centering
	\caption{Quantitative results of models with different window sizes for ×4 SR in which  Our-small is based. }
	\begin{tabular}{ccccccccc}
		\hline
		\multirow{2}{*}{Window size} &  \multicolumn{2}{c}{Set5} & \multicolumn{2}{c}{Set14} & \multicolumn{2}{c}{B100} & \multicolumn{2}{c}{Urban100}   \\
		& PSNR & SSIM              & PSNR & SSIM               & PSNR & SSIM              & PSNR & SSIM                                \\
		\hline
		(8,8)   & \textbf{32.406} & \textbf{0.8982} & \textbf{28.723} & \textbf{0.7856} &  \textbf{27.705} & \textbf{0.7415} & \textbf{26.435} & \textbf{0.7983} \\
		(16,16)  & 32.378 & 0.8973 & 28.735 & 0.7853 &  27.687 & 0.7409 & 26.340 & 0.7960  \\
		\hline
		\label{tab:window}
	\end{tabular}
\end{table*}

\begin{table*}
	\centering
	\caption{Quantitative results of models with different patch sizes for ×4 SR in which  Our-small is based. }
	\begin{tabular}{ccccccccc}
		\hline
		\multirow{2}{*}{Patch size} &  \multicolumn{2}{c}{Set5} & \multicolumn{2}{c}{Set14} & \multicolumn{2}{c}{B100} & \multicolumn{2}{c}{Urban100}  \\
		& PSNR & SSIM              & PSNR & SSIM               & PSNR & SSIM              & PSNR & SSIM                           \\
		\hline
		(48,48)  & \textbf{32.406} & \textbf{0.8982} & \textbf{28.723} & \textbf{0.7856} &  \textbf{27.705} & \textbf{0.7415} & \textbf{26.435} & \textbf{0.7983} \\
		(64,64)  & 32.286 & 0.8956 & 28.686 & 0.7846 &  27.653 & 0.7393 & 26.205 & 0.7917   \\
		\hline
		\label{tab:image}
	\end{tabular}
\end{table*}

\begin{figure}[t]
    \centering
    \vspace{-2mm}
    \begin{subfigure}{0.49\linewidth}
        \includegraphics[width=0.9\textwidth]{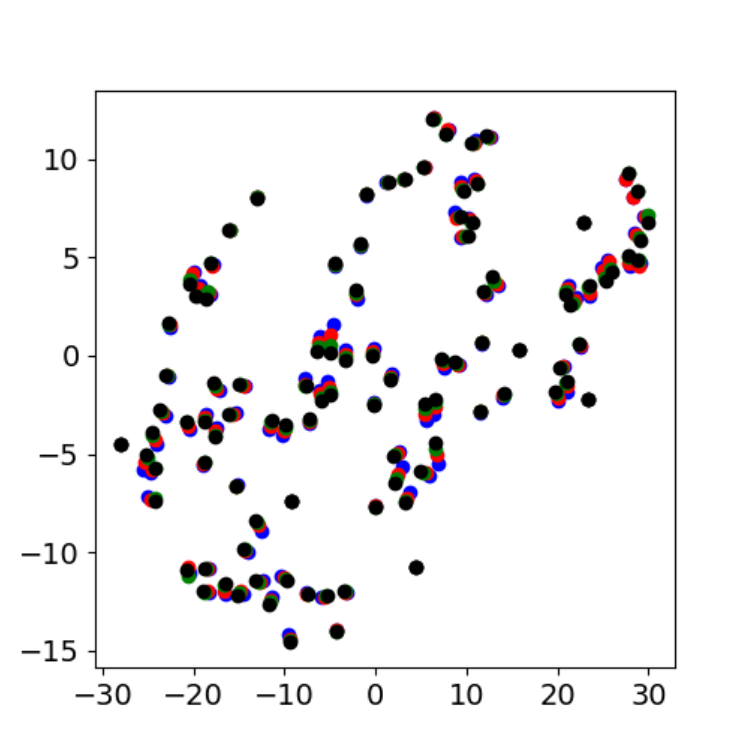}
        \caption{}
        \label{fig:3-a}
    \end{subfigure}
    \hfill
    \begin{subfigure}{0.49\linewidth}
        \includegraphics[width=0.9\textwidth]{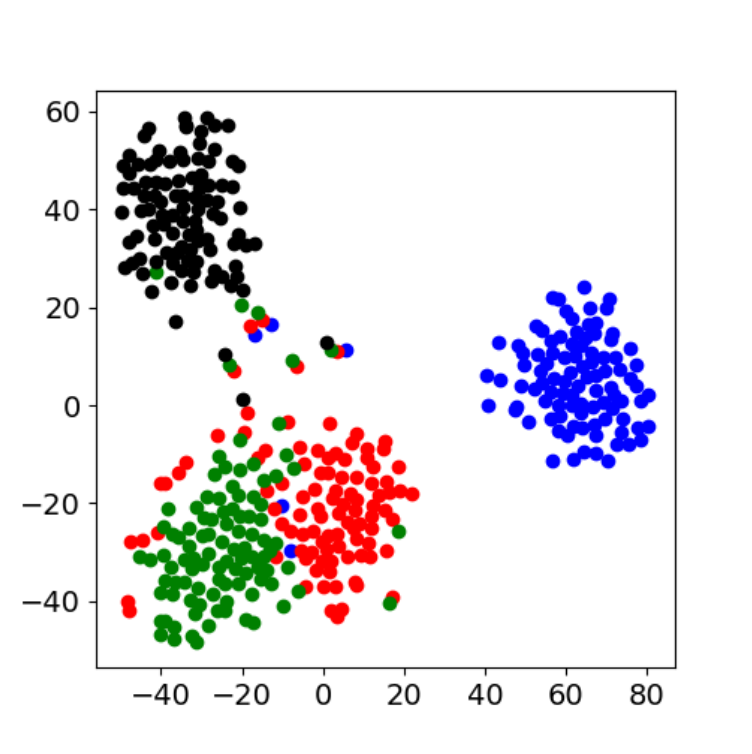}
        \caption{}
        \label{fig:3-b}
    \end{subfigure}
    \caption{Visualization of degradation characterization of $\sigma$ with different nuclear widths. (a) illustrates the representations generated by model 1 w/o degradation representation learning. (b) presents the representations generated by our model 5.}
    \label{fig:3}
\end{figure}

	The $query, key$ and $value$ matrices $Q, K$ and $V$ are computed as follows
	\begin{equation} 
		Q=X P_{Q}, \quad K=X P_{K}, \quad V=X P_{V} \odot D2
	\end{equation}
	where $P_{Q}$, $P_{K}$, and $P_{V}$ are projection matrices that are shared across different windows, $D2$ is the channel attention weights calculated from the learned degradation, and $\odot$ represents channel-wise multiplication. Generally, we have $Q, K, V \in \mathbb{R}^{M^{2}\times d}$. The attention matrix is thus computed by the self-attention mechanism in a local window as follows
	\begin{equation} 
		\operatorname{Attention}(Q, K, V)=\operatorname{SoftMax}(\dfrac{Q K^{T}}{\sqrt{d}}+B )V 
	\end{equation}
where $B$ is the learnable relative positional encoding. Then, we use a multi-layer perceptron (MLP) which has two fully-connected layers with GELU non-linearity between them for further feature transformation. The LayerNorm (LN) layer is added before both the MSA and MLP, and the residual connection is employed for both modules. The whole process is formulated as follows

\begin{equation} 
    X^{'}=\operatorname{DMSA}(\mathrm{LN}(X))+X
\end{equation}
\begin{equation} 
    X^{''}=\operatorname{MLP}(\mathrm{LN}(X^{'}))+X^{'}.
\end{equation}

\section{Experiments}

\subsection{Experimental Setup}
For image SR, the DRSTB number, CMT number, window size, channel number, and attention head number are set to 6, 6, $8\times8$, 180, and 6, respectively. We generate LR images according to~(\ref{eq:16}) for training and testing. We use 800 training images in DIV2K and 2650 training images in Flickr2K as the training set, and include four benchmark datasets (i.e., Set5, Set14, B100, and Urban100) for evaluation. The size of the Gaussian kernel is fixed to 21 × 21. We first train our method on noise-free degradation with isotropic Gaussian kernels only. The ranges of the kernel widths $\sigma$ are set to [0.2,2.0], [0.2,3.0], and [0.2,4.0] for $\times 2/3/4$ SR, respectively. Then, our method is trained on more general types of degradation with anisotropic Gaussian kernels and noise. Anisotropic Gaussian kernels characterized by a Gaussian probability density function $N(0, \sum)$ (with zero mean and varying covariance matrix $\sum$) are considered. The covariance matrix $\sum$ is determined by two random eigenvalues $\lambda_{1}, \lambda_{2} \thicksim U(0.2, 4)$ and a random rotation angle $\Theta \thicksim U(0, \pi)$. The noise level ranges from 0 to 25.

In training, we randomly select 16 HR images, and data augmentation is performed via random rotation and flipping. Next, we randomly select 16 Gaussian kernels to generate LR images. Then, we randomly crop 16 LR patches of size $48\times48$ (each LR image has two patches) and their corresponding HR patches as follows 
\begin{equation} 
	I^{L R}=\left(I^{H R} \otimes k\right) \downarrow_{s}+n
	\label{eq:16}
\end{equation}
where $I^{H R}$ is the HR image, $k$ is a blur kernel, $\otimes$ denotes the convolution operator, $ \downarrow_{s}$ represents the downsampling operator with a scale factor of $s$, and $n$ refers to the additive white Gaussian noise. We use the bicubic downsampler as the downsampling operator. 

\subsection{Implementation Details}
We adopt ADAM as the optimizer, and $\beta_1$ and $\beta_2$ are set to 0.9 and 0.999, respectively. We use L1 as the loss function with an initial learning rate of 2e-4, and decrease the learning rate by half every 250 epochs. We train a degradation learning module of 300 epochs, and the whole network is trained for 1000 epochs. Please refer to the open source code for more details of the experiments at: https://github.com/I2-Multimedia-Lab/DSAT/tree/main.
\begin{figure*}
	\centering
	\vspace{-1mm}
	\includegraphics[width=0.9 \linewidth]{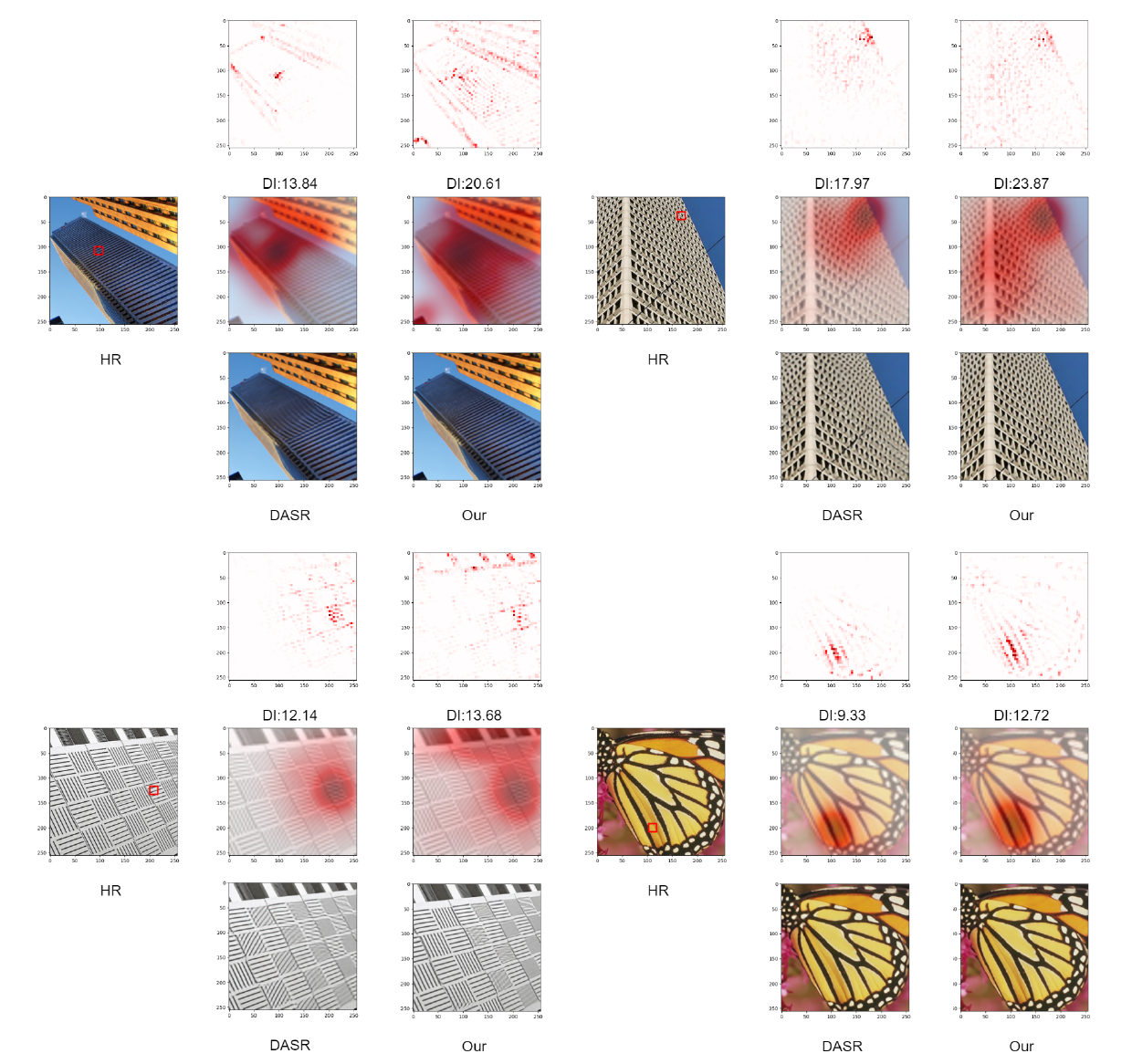}
	\caption{LAM comparisons between DASR and Ours. From top to bottom: the images are the LAM result, the informative area with the input image, and the super-resolved image.}
	\label{lam1}
\end{figure*}
\begin{figure}[t]
	\centering
	\begin{subfigure}{0.49\linewidth}
		\includegraphics[width=0.9\textwidth]{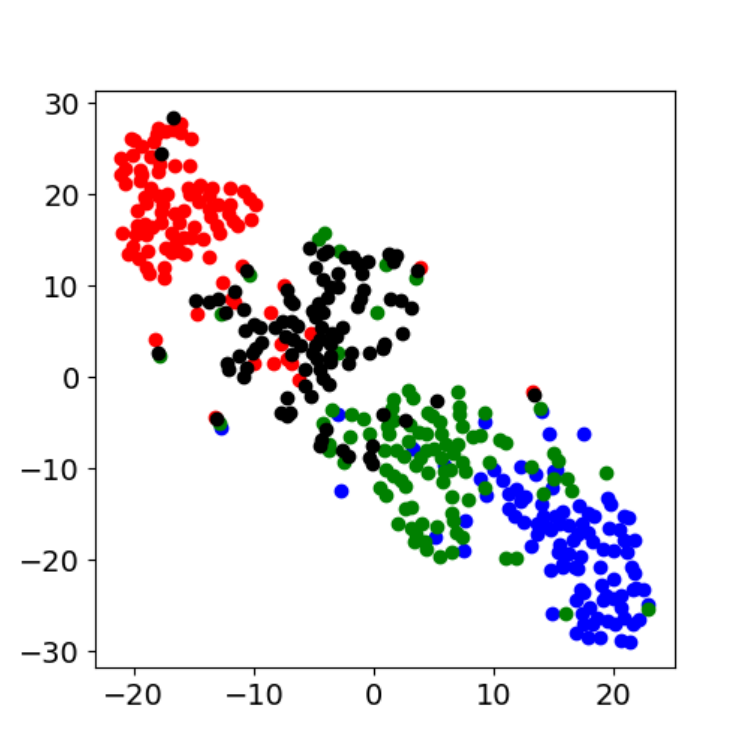}
		\caption{}
		\label{fig:5-a}
	\end{subfigure}
	\hfill
	\begin{subfigure}{0.49\linewidth}
		\includegraphics[width=0.9\textwidth]{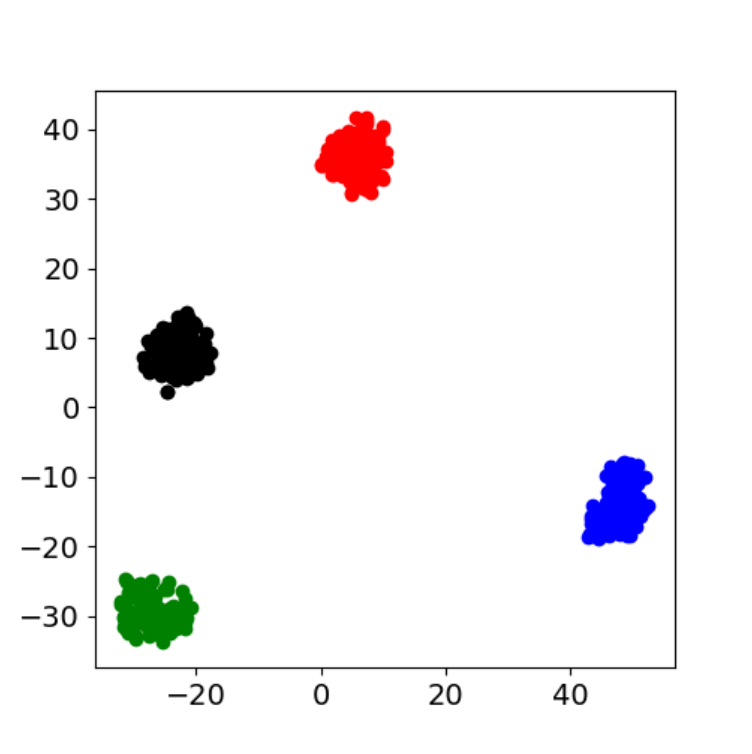}
		\caption{}
		\label{fig:5-b}
	\end{subfigure}
	\caption{Visualization of representations for degradation with: (a) different blur kernels and noise-free; and (b) the same blur kernel and different noise levels.}
	\label{fig:5}
	\vspace{-3mm}
\end{figure}

\begin{figure*}[t]
    \centering
    \includegraphics[width=1.0 \linewidth]{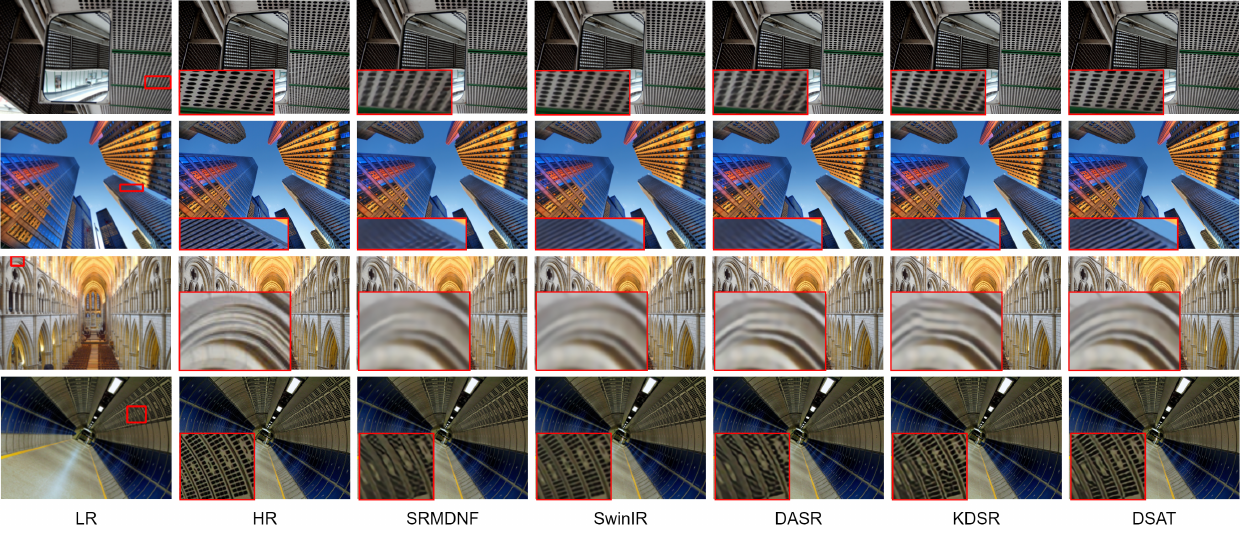}
    \caption{Visual comparisons achieved on Urban100 for ×4 SR.}
    \label{fig:4}
\end{figure*}
\subsection{Ablation Experiments}
We first perform ablation experiments on degradation with the isotropic Gaussian kernel and noise. We conduct separate quantitative and qualitative ablation experiments for our proposed method. Firstly, for the quantitative analysis, we evaluate the performance of each model using two metrics, i.e., PSNR and SSIM. We employ a rigorous experimental design to ensure the accuracy of the results. Specifically, Model 1 removes the contrastive learning module. Model 2 does not utilize degradation information in the attention mechanism. Model 3 does not incorporate the DCL in CMT. Model 4 removes both the DCL and degradation-aware attention weight (D2). Finally, Model 5 refers to our proposed network. As shown in Table \ref{table:1},
the results show that Model 5 significantly outperforms all the other models in all evaluation metrics.

In addition to the quantitative analysis, we also perform a qualitative analysis to assess the subjective quality of the results produced by each model. The results of the qualitative analysis, as shown in Fig. \ref{fig:alb}, clearly suggest that Model 5 produces the most visually pleasing results. In summary, both quantitative and qualitative analysis have proven that the degradation learning module can indeed produce better local details, while the DCL module and the Attention Weights module can make local details sharper. To ensure the completeness of the ablation experiments, we carry out additional ablation experiments on the patch size and window size. The results are shown in Tables \ref{tab:window} and \ref{tab:image}, respectively. Unlike other Transformer-based methods, increasing the patch and window sizes while keeping other settings unchanged does not improve the performance but rather makes it worse. We believe this phenomenon may be caused by the degradation information injected into the SR network.

\subsection{Experiments on Noise-free Degradation with Isotropic Gaussian Kernels}

We first conduct experiments on isotropic Gaussian kernels in the absence of noise. We qualitatively and quantitatively analyze how degradation representation learning works, and then compare our model with its state-of-the-art counterparts under a blind setting.

	\begin{table*}[t]
		\caption{PSNR$\uparrow$ results achieved on noise-free degradation with isotropic Gaussian kernels. Note that the degradation becomes bicubic degradation when the kernel width is set to 0. The best and second best performers are marked in red and blue, respectively.}
		\centering
		\label{table:2}
		\resizebox{\textwidth}{!}{
			\begin{tabular}{cccccccccccccccccc}
				\hline
				Method&
				\multirow{2}{*}{Scale}&
				\multicolumn{4}{c}{Set5}& \multicolumn{4}{c}{Set14}& \multicolumn{4}{c}{B100}& \multicolumn{4}{c}{Urban100} \\
				Kernel Width & &  0   &  0.6  &  1.2  &  1.8  &   0   &  0.6  &  1.2  &  1.8  &   0   &  0.6  &  1.2  &  1.8  &   0   &  0.6  &  1.2  &  1.8   \\
				\hline
				Bicubic & \multirow{8}{*}{$\times$2}      & 33.66 & 32.30 & 29.28 & 27.07 & 30.24 & 29.21 & 27.13 & 25.47 & 29.56 & 28.76 & 26.93 & 25.51 & 26.88 & 26.13 & 24.46 & 23.06  \\
				RCAN \cite{zhang2018image}&              & {\color{blue}38.27} & 35.91 & red.20 & 28.50 & {\color{blue}34.12} & 32.31 & 28.48 & 26.33 & {\color{blue}32.41} & 31.16 & 28.04 & 26.26 & {\color{blue}33.34} & 29.80 & 25.38 & 23.44  \\
				SRMDNF \cite{zhang2018learning}  + Predictor \cite{gu2019blind} &              & 34.94 & 34.77 & 34.13 & 33.80 & 31.48 & 31.35 & 30.78 & 30.18 & 30.77 & 30.33 & 29.89 & 29.20 & 29.05 & 28.42 & 27.43 & 27.12  \\
				MZSR \cite{soh2020meta}  + Predictor \cite{gu2019blind}   &              & 35.96 & 35.66 & 35.22 & 32.32 & 31.97 & 31.33 & 30.85 & 29.17 & 30.64 & 29.82 & 29.41 & 28.72 & 29.49 & 29.01 & 28.43 & 26.39  \\
				SwinIR \cite{liang2021swinir}                &              & ${\color{red}38.37}$&$35.96$&$31.21$&$28.51$&${\color{red}34.14}$&$32.38$&$28.49$&$26.33$&${\color{red}32.46}$&$31.19$&$28.04$&$26.26$&${\color{red}33.43}$&$29.92$&$25.39$&$23.45$ \\
				DASR \cite{wang2021unsupervised}               &              & 
				37.87 & {\color{blue}37.47} &{\color{blue} 37.19} &{\color{blue} 35.43} & 
				33.34 &{\color{blue}32.96} &{\color{blue} 32.78} &{\color{blue}31.60} & 
				32.03 & {\color{blue}31.78} & {\color{blue}31.71} &{\color{blue} 30.54} &
				31.49 & {\color{blue}30.71} & {\color{blue}30.36 }& {\color{blue}28.95}  \\
				DSAT                &              &  
				38.14 & {\color{red}38.06} & {\color{red}37.59} & {\color{red}35.65} & 
				33.81 & {\color{red}33.55} & {\color{red}33.34} & {\color{red}31.88} & 
				32.29 & {\color{red}32.21} & {\color{red}32.10} & {\color{red}30.88} & 
				32.43 & {\color{red}31.50} & {\color{red}31.03} & {\color{red}29.40}  \\ 
				
				\hline
				Kernel Width & &   0   &  0.8  &  1.6  &  2.4  &   0   &  0.8  &  1.6  &  2.4  &   0   &  0.8  &  1.6  &  2.4  &   0   &  0.8  &  1.6  &  2.4   \cr\hline
				Bicubic & \multirow{5}{*}{$\times$3}      & 30.39 & 29.42 & 27.24 & 25.37 & 27.55 & 26.84 & 25.42 & 24.09 & 27.21 & 26.72 & 25.52 & 24.41 & 24.46 & 24.02 & 22.95 & 21.89  \\
				RCAN \cite{zhang2018image}  &              & {\color{blue}34.74} & 32.90 & 29.12 & 26.75 & {\color{blue}30.65} & 29.49 & 26.75 & 24.99 & {\color{blue}29.32} & 28.56 & 26.55 & 25.18 & {\color{blue}29.09} & 26.89 & 26.89 & 22.30  \\
				SRMDNF \cite{zhang2018learning}  + Predictor \cite{gu2019blind} &              & 32.22 & 32.63 & 32.27 & 28.62 & 29.13 & 29.25 & 28.01 & 26.90 & 28.41 & 28.25 & 28.11 & 26.56 & 26.75 & 26.61 & 26.35 & 24.06  \\
				SwinIR \cite{liang2021swinir}               &              &${\color{red}34.89}$&$32.98$&$29.12$&$26.76$&${\color{red}30.77}$&$29.59$&$26.77$&$25.00$&${\color{red}29.38}$&$28.62$&$26.56$&$25.18$&${\color{red}29.30}$&$27.05$&$23.86$&$22.30$ \\
				DASR \cite{wang2021unsupervised}               &              &
				24.06 &{\color{blue}34.08} & {\color{blue}33.57} &{\color{blue} 32.15} & 
				30.13 &{\color{blue} 29.99 }& {\color{blue}28.66 }& {\color{blue}28.42} & 
				28.96 & {\color{blue}28.90 }& {\color{blue}28.62 }& {\color{blue}28.13 }&
				27.65 & {\color{blue}27.36} &{\color{blue} 26.86 }& {\color{blue}25.95} \\
				DSAT                &              & 
				34.66 & {\color{red}34.56} & {\color{red}33.77} & {\color{red}31.96} & 
				30.59 & {\color{red}30.48} & {\color{red}30.17} & {\color{red}28.98} & 
				29.26 & {\color{red}29.22} & {\color{red}29.10} & {\color{red}28.24} & 
				28.73 & {\color{red}28.38} & {\color{red}27.88} & {\color{red}26.67}  \\
				\hline
				Kernel Width & &   0   &  1.2  &  2.4  &  3.6  &   0   &  1.2  &  2.4  &  3.6  &   0   &  1.2  &  2.4  &  3.6 &   0   &  1.2  &  2.4  &  3.6    \cr\hline
				Bicubic & \multirow{6}{*}{$\times$4}      & 28.42 & 27.30 & 25.12 & 23.40 & 26.00 & 25.24 & 23.83 & 22.57 & 25.96 & 25.42 & 24.20 & 23.15 & 23.14 & 22.68 & 21.62 & 20.65  \\
				RCAN \cite{zhang2018image} &              & {\color{blue}32.63} & 30.26 & 26.72 & 24.66 & {\color{blue}28.87} & 27.48 & 24.93 & 23.41 & 27.72 & 26.89 & 25.09 & 23.93 & 26.61 & 24.71 & 22.25 & 20.99  \\
				SRMDNF \cite{zhang2018learning}  + Predictor \cite{gu2019blind} &              & 30.61 & 29.35 & 29.27 & 28.65 & 27.74 & 26.15 & 26.20 & 26.17 & 27.15 & 26.15 & 26.15 & 26.14 & 25.06 & 24.11 & 24.10 & 24.08  \\
				IKC \cite{gu2019blind}                 &              & 32.00 & 31.77 & 30.56 & 29.23 & 28.52 & {\color{blue}28.45} & 28.16 & 26.81 & 27.51 & 27.43 & 27.27 & 26.33 & 25.93 & 25.63 & 25.00 & 24.06  \\
				SwinIR \cite{liang2021swinir}                 &              &${\color{red}32.76}$&$30.35$&$26.73$&$24.67$&${\color{red}28.94}$&$27.54$&$24.94$&$23.42$&${\color{red}27.85}$&$26.92$&$25.10$&$23.94$&${\color{red}27.08}$&$24.82$&$22.27$&$20.99$ \\
				DASR \cite{wang2021unsupervised} &              & 31.99 & 31.92 & 31.75 & {\color{blue}30.59} & 28.50 & 28.45 & 28.28 & 27.45 & 27.51 & 27.52 & 27.43 & 26.83 & 25.82 &25.69 & 25.44 & 24.66  \\
				KDSR \cite{xia2022knowledge} &              &
				32.42 &  {\color{blue}32.34} &  {\color{red}32.13} &  {\color{red}31.02} &
				28.67 &  {\color{blue}28.66} &  {\color{red}28.55} &  {\color{red}27.81} &
				27.64 &  {\color{blue}27.67} &  {\color{blue}27.60} &  {\color{blue}26.98} &
				26.36 &  {\color{blue}26.29} &  {\color{red}26.06} &  {\color{red}25.21}  \\
				DSAT                &              &
				32.52&{\color{red}32.51}&{\color{blue}32.00}&30.31&
				28.85& {\color{red}28.77} & {\color{blue}28.50} & {\color{blue}27.51}&
				{\color{blue}27.76}& {\color{red}27.76} & {\color{red}27.66} & {\color{red}27.02}&
				{\color{blue}26.62}& {\color{red}26.43} & {\color{blue}25.95} & {\color{blue}24.89} \\
				\hline
			\end{tabular}
		}
	\end{table*}

\begin{table*}[t]
	\caption{PSNR$\uparrow$ results achieved on Set14 for ×4 SR with anisotropic Gaussian kernels and noises. The best and second best performers are marked in red and blue, respectively.}
	\label{table:3}
	\centering
	\resizebox{\textwidth}{!}{
		\begin{tabular}{ccc c c c c c c c c}
			\hline
			\multirow{3}{*}{method}                   & \multirow{3}{*}{noise} & \multicolumn{9}{c}{Blur Kernel}\\
			&&\includegraphics[height=20pt,width=20pt]{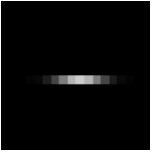}
			&\includegraphics[height=20pt,width=20pt]{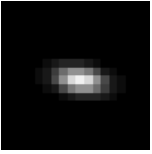}
			&\includegraphics[height=20pt,width=20pt]{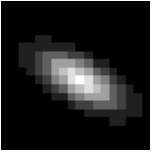}
			&\includegraphics[height=20pt,width=20pt]{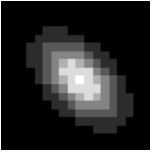}
			&\includegraphics[height=20pt,width=20pt]{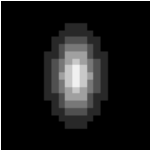}
			&\includegraphics[height=20pt,width=20pt]{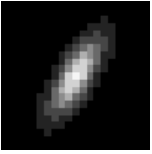}
			&\includegraphics[height=20pt,width=20pt]{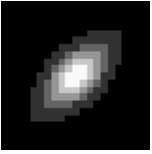} 
			&\includegraphics[height=20pt,width=20pt]{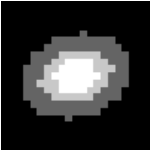} 
			&\includegraphics[height=20pt,width=20pt]{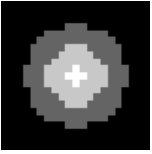}        \\
			\hline
			
			\multirow{3}{*}{DnCNN+RCAN}               & 0                      & 26.44 & 26.22 & 24.48 & 24.23 & 24.29 & 24.19 & 23.9  & 23.42 & 23.01  \\
			& 5                      & 26.10 & 25.90 & 24.29 & 24.07 & 24.14 & 24.02 & 23.74 & 23.31 & 22.92  \\
			& 10                     & 25.65 & 25.47 & 24.05 & 23.84 & 23.92 & 23.8  & 23.54 & 23.14 & 22.77  \\\hline
			\multirow{3}{*}{DnCNN +SRMDNF +Predictor} & 0                      & 26.84 & 26.88 & 25.57 & 25.69 & 25.64 & 24.98 & 25.12 & 25.28 & 24.84  \\
			& 5                      & 25.92 & 25.75 & 24.18 & 23.97 & 24.05 & 23.93 & 23.65 & 23.20 & 22.80  \\
			& 10                     & 25.39 & 25.23 & 23.88 & 23.68 & 23.74 & 23.65 & 23.39 & 22.99 & 22.64  \\\hline
			\multirow{3}{*}{DnCNN +IKC}               & 0                      & 27.71 & 27.78 & 27.11 & 27.02 & 26.93 & 26.65 & 26.5  & 26.01 & 25.33  \\
			& 5                      & 26.91 & 26.80 & 24.87 & 24.53 & 24.56 & 24.40 & 24.06 & 23.53 & 23.06  \\
			& 10                     & 26.16 & 26.09 & 24.55 & 24.33 & 24.35 & 24.17 & 23.92 & 23.43 & 23.01  \\\hline
			\multirow{3}{*}{DnCNN + DCLS \cite{luo2022deep}}                      & 0                      & 27.56 & 27.49 & 26.32 & 25.99 & 25.88 & 26.03 & 25.70 & 24.65 & 23.95  \\
			& 5                      & 26.20 & 26.02 & 24.44 & 24.21 & 24.28 & 24.14 & 23.88 & 23.40 &  22.98  \\
			& 10                     & 25.47 & 25.33 & 24.06 & 23.87 & 23.91 & 23.79 & 23.58 & 23.16 & 22.78  \\\hline
			\multirow{3}{*}{DASR}                     & 0                      & {\color{blue}27.99} & {\color{blue}27.97} & {\color{blue}27.53} & {\color{blue}27.45}& {\color{blue}27.43} & {\color{blue}27.22} & {\color{blue}27.19} & {\color{blue}26.83} & {\color{blue}26.21}  \\
			& 5                      & {\color{blue}27.25} & {\color{blue}27.18} & {\color{blue}26.37} & {\color{blue}26.16} & {\color{blue}26.09} & {\color{blue}25.96} & {\color{blue}25.85} & {\color{blue}25.52} & {\color{blue}25.04}  \\
			& 10                     & {\color{blue}26.57} & {\color{blue}26.51} & {\color{blue}25.64} & {\color{blue}25.47} & {\color{blue}25.43} & {\color{blue}25.31} & {\color{blue}25.16} & {\color{blue}24.80} & {\color{blue}24.43}  \\\hline
			\multirow{3}{*}{DSAT}                     & 0                      & {\color{red}28.34} & {\color{red}28.34} & {\color{red}27.78} & {\color{red}27.68} & {\color{red}27.68} & {\color{red}27.37} & {\color{red}27.25} & {\color{red}26.98} &{\color{red}26.47}  \\
			& 5                      & {\color{red}27.55} & {\color{red}27.47} & {\color{red}26.59} & {\color{red}26.43} & {\color{red}26.43} & {\color{red}26.31} & {\color{red}26.14} & {\color{red}25.80} & {\color{red}25.36}  \\
			& 10                     & {\color{red}26.83} & {\color{red}26.74} & {\color{red}25.87} & {\color{red}25.71} & {\color{red}25.71} & {\color{red}25.60}  & {\color{red}25.44} & {\color{red}25.10} & {\color{red}24.70}  \\\hline
	\end{tabular}}
	\vspace{-2mm}
\end{table*}

\begin{table*}
	\caption{More results on the dataset Manga109.}
	\centering
	\label{table:aa}
	\begin{tabular}{cccccccccc}
		\hline
		Kernel Width & & \multicolumn{2}{c}{0}   &  \multicolumn{2}{c}{0.6}  & \multicolumn{2}{c}{ 1.2}  &  \multicolumn{2}{c}{1.8}    \\
		Method& &  PSNR $\uparrow$ &  SSIM $\uparrow$ &  PSNR $\uparrow$  &  SSIM $\uparrow$  &  PSNR $\uparrow$  &  SSIM $\uparrow$&  PSNR $\uparrow$  &  SSIM $\uparrow$ \\
		\hline
		DASR       &       \multirow{2}{*}{$\times$2}        & 
		38.018 & 0.9763 & 37.363 & 0.9734 &  36.339 & 0.9701 & 34.751 & 0.9557 \\
		DSAT       &              &  
		\textbf{39.016} & \textbf{0.9781} & \textbf{38.662} & \textbf{0.9766} & \textbf{37.837} & \textbf{0.9728} & \textbf{34.955} & \textbf{0.9562} \\ 
		
		\hline
		\multirow{2}{*}{Kernel Width} & &  \multicolumn{2}{c}{ 0}   &  \multicolumn{2}{c}{0.8}  &  \multicolumn{2}{c}{1.6}  & \multicolumn{2}{c}{ 2.4} \\
		& &   PSNR $\uparrow$ &  SSIM $\uparrow$ &  PSNR $\uparrow$  &  SSIM $\uparrow$  &  PSNR $\uparrow$  &  SSIM $\uparrow$&  PSNR $\uparrow$  &  SSIM $\uparrow$ \cr\hline
		DASR        &        \multirow{2}{*}{$\times$3}       &
		32.948 & 0.9418 & 33.028 & 0.9385 & 32.576 &  0.9354 & 30.968 & 0.9153 \\
		DSAT         &              & 
		\textbf{33.945} & \textbf{0.9486} & \textbf{34.026} & \textbf{0.9470} & \textbf{33.601} & \textbf{0.9432} & \textbf{31.921} & \textbf{0.9260} \\
		\hline
		\multirow{2}{*}{Kernel Width} & &   \multicolumn{2}{c}{0} &  \multicolumn{2}{c}{0.4}  & \multicolumn{2}{c}{1.2}  &  \multicolumn{2}{c}{2.0}  \\
		& &  PSNR $\uparrow$ &  SSIM $\uparrow$ &  PSNR $\uparrow$  &  SSIM $\uparrow$  &  PSNR $\uparrow$  &  SSIM $\uparrow$&  PSNR $\uparrow$  &  SSIM $\uparrow$ \cr\hline
		DASR     &   \multirow{3}{*}{$\times$4} &
		30.164 & 0.9057 & 30.211 & 0.9056 & 30.155&0.9016 & 30.132 & 0.8998 \\
		KDSR   &         &
		30.812 &  0.9112 &  30.818 & 0.9110 & 30.910 & 0.9100 & \textbf{30.854} & 0.9076  \\
		DSAT      &              &
		\textbf{30.970} & \textbf{0.9167} & \textbf{31.017} & \textbf{0.9167} & \textbf{31.041} & \textbf{0.9153} & 30.769 & \textbf{0.9115} \\
		\hline
	\end{tabular}
\end{table*}

\begin{figure*}[t]
	\centering
	\begin{subfigure}{0.25\linewidth}
		\includegraphics[width=120pt]{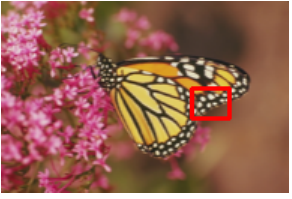}
	\end{subfigure}
	\begin{subfigure}{0.25\linewidth}
		\includegraphics[width=120pt]{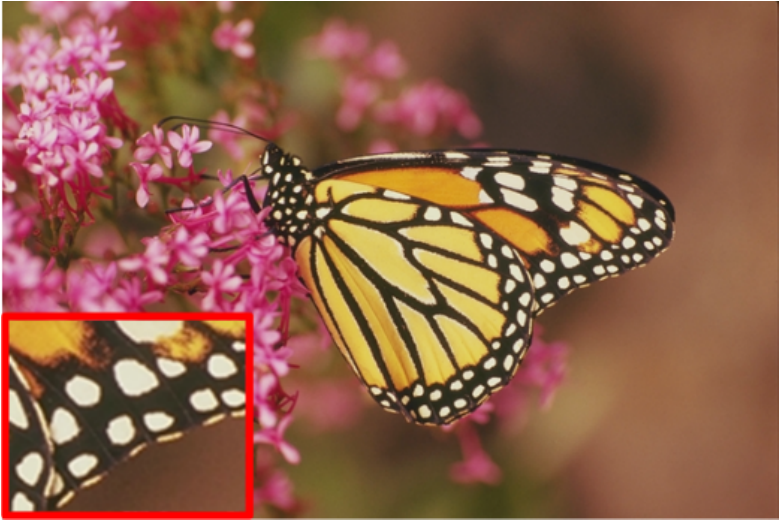}
	\end{subfigure}
	\begin{subfigure}{0.25\linewidth}
		\includegraphics[width=120pt]{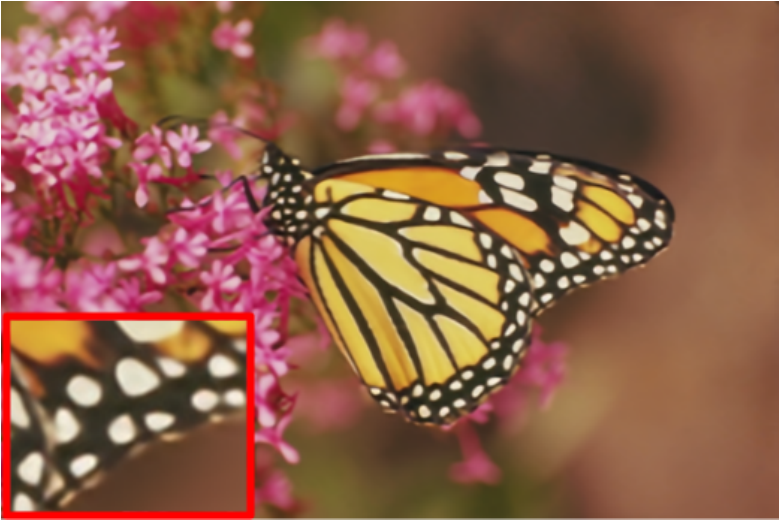}
	\end{subfigure}
	\begin{subfigure}{0.20\linewidth}
		\includegraphics[width=120pt]{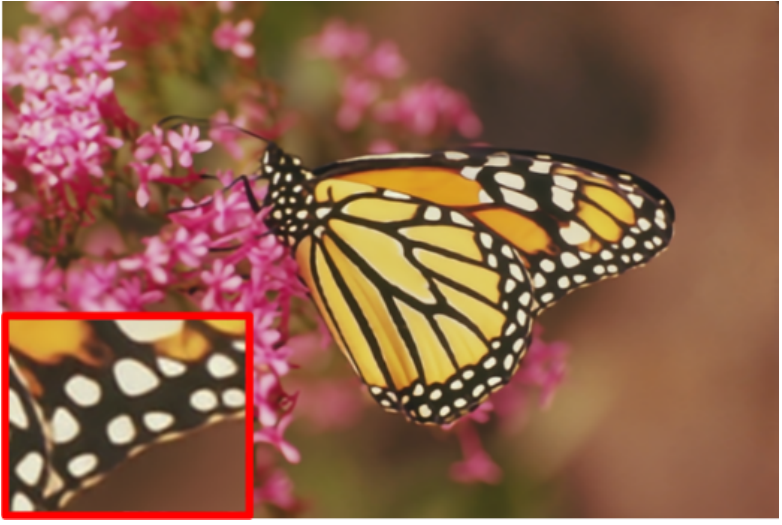}
	\end{subfigure}
	
	\begin{subfigure}{0.25\linewidth}
		\includegraphics[width=120pt]{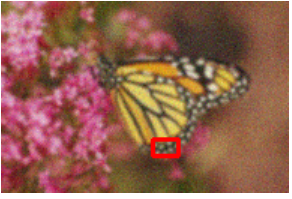}
	\end{subfigure}
	\begin{subfigure}{0.25\linewidth}
		\includegraphics[width=120pt]{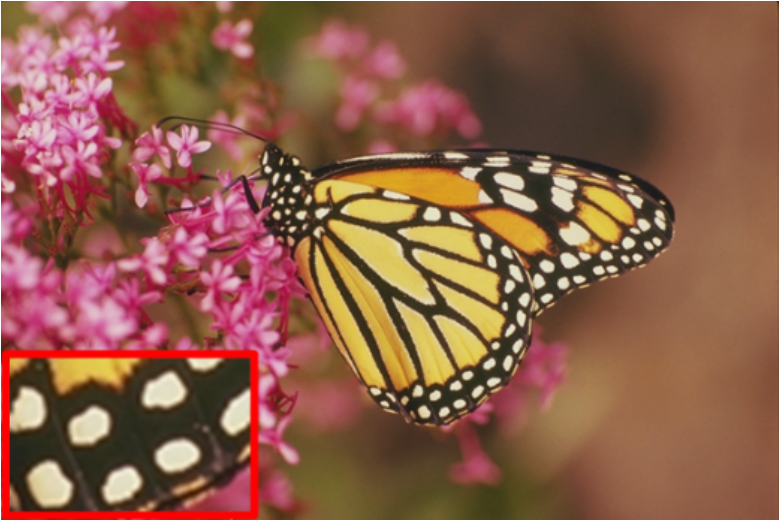}
	\end{subfigure}
	\begin{subfigure}{0.25\linewidth}
		\includegraphics[width=120pt]{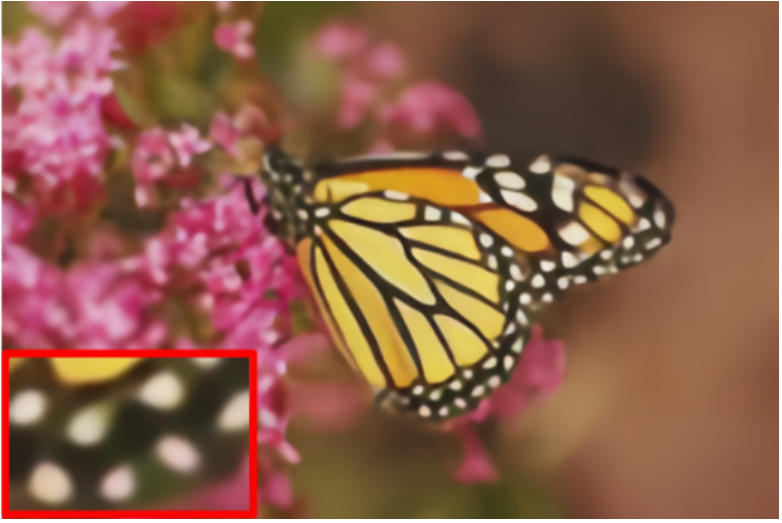}
	\end{subfigure}
	\begin{subfigure}{0.20\linewidth}
		\includegraphics[width=120pt]{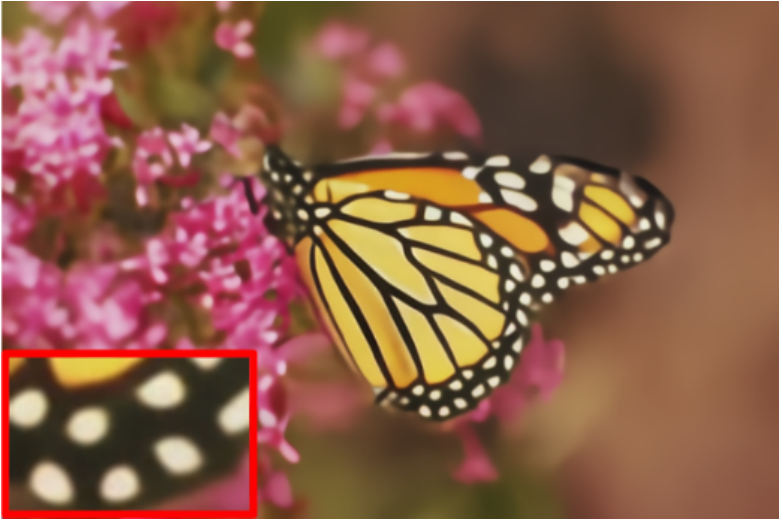}
	\end{subfigure}
	
	\begin{subfigure}{0.25\linewidth}
		\raisebox{-1.01\height}{\includegraphics[width=120pt]{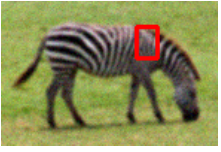}}
		\caption*{LR image}
	\end{subfigure}
	\begin{subfigure}{0.25\linewidth}
		\includegraphics[width=120pt]{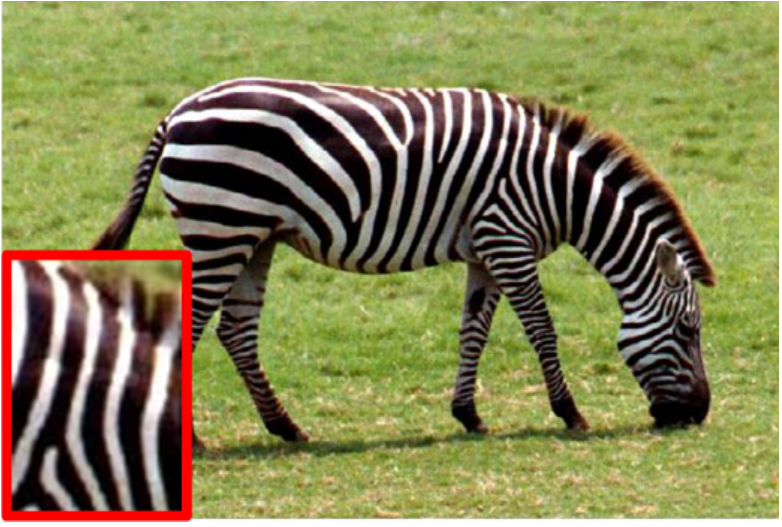}
		\caption*{GT}
	\end{subfigure}
	\begin{subfigure}{0.25\linewidth}
		\includegraphics[width=120pt]{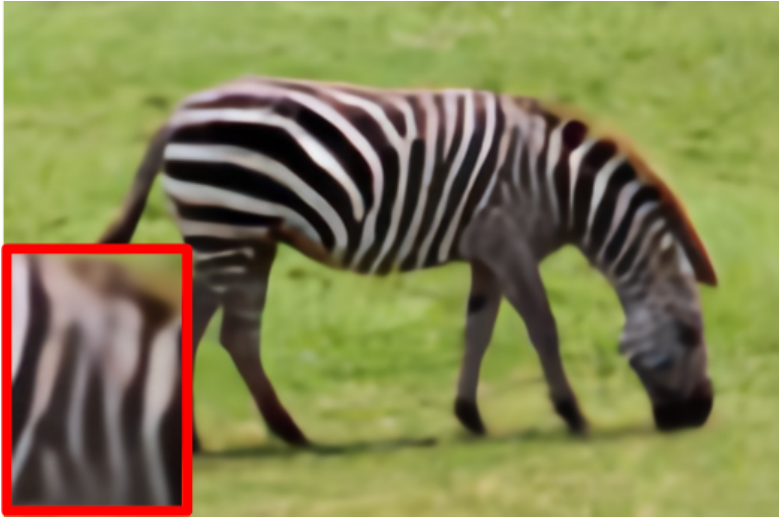}
		\caption*{DASR}
	\end{subfigure}
	\begin{subfigure}{0.20\linewidth}
		\includegraphics[width=120pt]{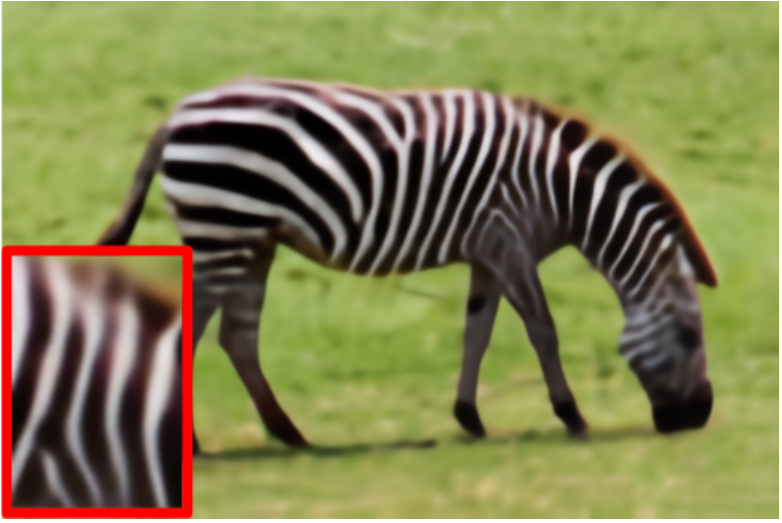}
		\caption*{DSAT}
	\end{subfigure}
	\caption{Visual comparisons achieved on Set14. The first row uses the first Blur Kernel in Table \ref{table:3}, the second and third rows use the ninth Blur Kernel in Table \ref{table:3}, and their noise levels are set to 0 and 10, respectively.}
	\label{fig:6}
\end{figure*}

\textbf{Degradation Representation Learning. }
We use different patches from the same LR image with complex degradation as positive samples, and different patches from other LR images as negative samples. These samples are used as inputs to contrastive learning in an attempt to maximize the similarity between positive samples and to minimize the similarity between positive and negative samples. The degradation from the positive samples are pulled closer together than the degradation from negative samples. The learned degradation representation is further used to guide the subsequent SR network for improving performance.

The purpose of degradation representation learning is to allow each image to produce different degradation information to assist the deep extraction module. As previously shown in Table \ref{table:1} and Fig. \ref{fig:alb}, the degradation representation learning does lead to better reconstruction results.

In addition, we undertake additional experiments to demonstrate the effectiveness of degradation representation learning. Specifically, we compare and visualize the degradation representations learned by Model 1 and Model 5. We first generate LR images with different types of degradation using the B100 dataset, then use Model 1 and Model 5 to generate degradation representations, and finally visualize the obtained degradation representations using the T-SNE method. As can be seen from Fig. \ref{fig:3-b}, our proposed model can learn and generate clustering results that are easier to discriminate. If degradation representation learning is removed, images with different degrees of degradation cannot be distinguished, and therefore no discriminative clusters can be obtained, as shown in Fig. \ref{fig:3-a}. This suggests that degradation representation learning can indeed obtain more discriminative degradation information, which helps our SR network discriminate LR images to produce more visually pleasant SR results. In addition, we compare the SR performances of Model 1 and Model 5 in Table \ref{table:1}. If degradation representation learning is removed, Model 1 does not adapt well to LR images with different degrees of degradation, especially when the kernel width becomes larger. However, our proposed can obtain more accurate degradation information for different kernel widths thanks to the degradation representation learning. As a result, the PSNRs achived by our proposed model are significantly better than those of model 1.

\textbf{Local Attribution Maps. }
We conduct an interpretability analysis of our results using the attribution analysis method LAM \cite{gu2021interpreting} and compare it with DASR. The LAM adopts the integral gradient method to explore which input pixels contribute most to the final performance. The red dots in the LAM results indicate the pixels used in reconstructing the patches marked with red boxes in the HR images. We then calculate the Diffusion Index (DI) to analyze the pixel range that the SR model can use. The more pixels used in reconstructing a specific input patch, the wider the distribution of red dots in the LAM and the higher the DI. As can be observed from Fig.~\ref{fig:lam}, our model has a larger range of pixel usage and yields a better reconstruction performance. Based on the LAM, we can validate that our model can acquire more pixel information to improve the SR performance of the model.

We provide a more qualitative and quantitative comparison between DASR and our method in terms of LAM. As shown in Fig. \ref{lam1}, our method can use a larger range of pixels to reconstruct the local area, while DASR can only use pixels in a local range. Thus the DI of our model is significantly higher than that of DASR. In terms of PSNR, our method generates SR results with higher PSNRs and thus better visual quality.

\textbf{Comparison to Previous Networks. }
Table \ref{table:2} compares our method to several recent SR methods, including the DASR \cite{wang2021unsupervised}, RCAN \cite{zhang2018image}, SRMD \cite{zhang2018learning}, MZSR \cite{soh2020meta}, IKC \cite{gu2019blind}, and KDSR \cite{xia2022knowledge}. As can be seen from the table, our method achieves a much better performance on almost all four benchmark datasets for all scale factors and kernel widths. In general, our method achieves a gain of 0.3-0.7 dB in PSNR, which demonstrates the effectiveness of our model. In addition, we test the SwinIR using images with unknown degradation and find its performance drops significantly. By contrast, the PSNR of our model does not decay significantly in response to the Gaussian kernels with larger kernel widths, indicating that our method is insensitive to degradation estimation errors, and has a better generalization ability. Therefore, our model is more applicable to a wide variety of degradation in the real world.
Even compared to the state-of-the-art KDSR, our proposed DSAT still achieves a better performance with most kernel widths on various datasets. Note that the KDSR employs knowledge distillation to estimate degradation, which may not be stable since it is difficult to guarantee that the student network has the same degradation extraction ability as the teacher network in practice. In addition, it should be noted that the KDSR only gives pre-trained models at the $\times$4 scale, so we only compare with the KDSR at that scale. 

\begin{figure*}
	\centering
	\vspace{-2mm}
	\includegraphics[width=0.8 \linewidth]{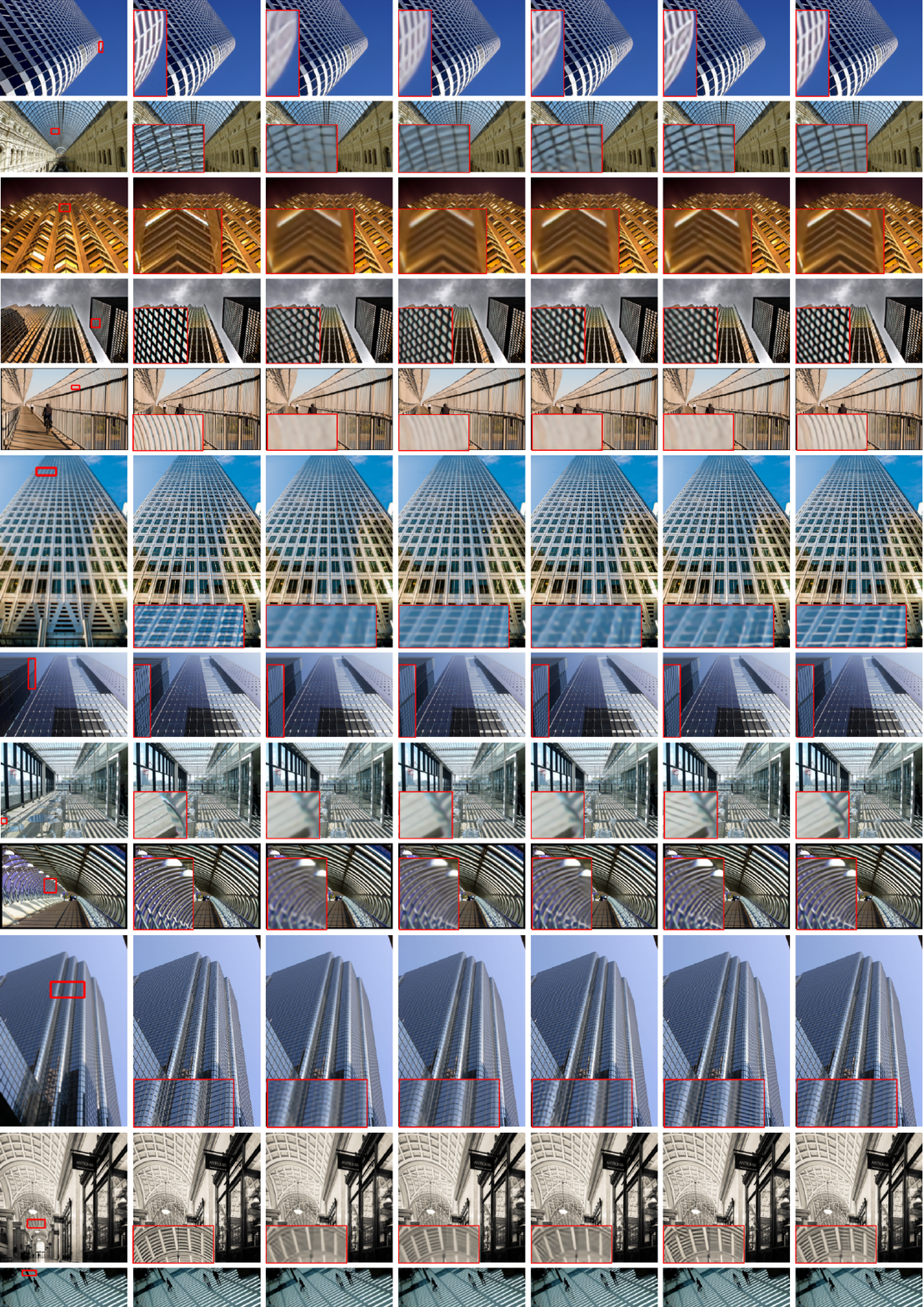}
	\caption{Visual comparisons of our proposed method with various state-of-the-art methods on the Urban100 which kernel width set to 1.2.}
	\label{suurban100}
\end{figure*}

We compare the visualization results on the Urban100 dataset. As shown in Figs.~\ref{fig:4} and~\ref{suurban100}, it is easy to see that our model has a better performance in preserving details and texture. In addition, our model also performs better in reconstructing graphical objects, fonts, and colors, on the Manga109 dataset, and more results are given in Supplementary Material.

\subsection{Experiments on General Degradation with Anisotropic Gaussian Kernels and Noise}
We further perform experiments on quality degradation with anisotropic Gaussian kernels and noise. We first analyze the learned general degradation and then compare our model with the state-of-the-art under a blind setting.

\textbf{Study of Degradation Representations. }
We conduct experiments to analyze the effect of the fuzzy kernel and noise on the degradation representation of the model. We first show the noise-free degradation representations of various fuzzy kernels in Fig. \ref{fig:5-a}. Then, we select one of the fuzzy kernels and show the clustering under different noise levels in Fig. \ref{fig:5-b}. As can be seen from the figures, our model can easily cluster degradation with different noise levels and roughly distinguish different fuzzy kernels.

\textbf{Comparison with Previous Networks. }
We use nine different fuzzy kernels and noise levels for performance evaluation. For a fair comparison, the fuzzy kernels we use are the same as in the DASR, so the settings of the RCAN, SRMDNF, IKC, and DCLS \cite{luo2022deep} are also the same.
As can be seen from Table \ref{table:3}, the RCAN does not perform well under complex degradation, while the SRMDNF is sensitive to degradation estimation errors. The IKC with iterative correction estimation performs significantly better than the SRMDNF. However, the IKC is more time-consuming due to the need for multiple iterations. The DASR can distinguish various types of degradation by learning discriminative representations. The CNN-based method DASR, while outperforming the above methods in all aspects, still suffers from a bottleneck in global modeling. Unlike the DASR, we use a CNN Mixed Transformer to efficiently utilize the obtained degenerate representation, and the results demonstrate that our method outperforms DASR in terms of PSNR for various fuzzy kernels and noise levels.
As shown in Fig. \ref{fig:6}, the first row visualizes the results of the models tested on the Set14 dataset with the first Blur Kernel and noise-free, while the second and third rows visualize the results of the models with the ninth Blur Kernel and noise $n$ being set to 0 and 10, respectively. We can clearly see that our model outperforms the DASR with and without noise. In summary, our model performs much better than the DASR under the anisotropic Gaussian kernel with different kernel widths and rotation angles.

\subsection{More Dataset Comparison}
To further demonstrate the superior performance of our model in reconstructing images, we conduct both qualitative and quantitative analysis on the Manga109 dataset. As shown in Table \ref{table:aa}, our model performs exceptionally well with all kernel widths at scales $\times$2,  $\times$3, and  $\times$4, and even outperforms the state-of-the-art KDSR model at most kernel widths. In addition, by setting the kernel width to 1.2 and conducting a qualitative analysis with our DSAT model and current mainstream methods, the results are shown in supplementary material. In summary, through qualitative and quantitative analyses, our model demonstrates a better performance, especially in the accuracy of texture reconstruction. Unlike the DASR and KDSR that wrongly reconstruct patterns in an image (e.g., the sharp lines on a building), our model does not produce erroneous texture, even though the DASR also uses degradation representation information via metric learning. Since the DASR uses the CNN as the backbone, this further demonstrates the effectiveness of using degradation representation information in guiding the learning of the Transformer for image super-resolution with unknown degradation.

\section{Conclusion}
In this paper, we proposed a degradation-based Transformer for blind image super-resolution to deal with various types of degradation. We considered implicit spatial degradation and employed contrastive learning to obtain discriminative features to distinguish various types of degradation. Our SR network specifically uses a CNN Mixed Transformer model to extract features, where we incorporate the learned degradation to make the network adaptive to specific degradation. Experimental results were presented to demonstrate that our degradation representation learning scheme can extract discriminative representations to obtain accurate degradation information. The experimental results also showed that our network yields the state-of-the-art blind SR performance under different degrees of degradation.


%




\ifCLASSOPTIONcaptionsoff
  \newpage
\fi



%

\bibliographystyle{ieee}
\bibliography{ref}

\end{document}